\documentclass[10pt,twocolumn,letterpaper]{article}

\usepackage{cvpr}
\usepackage{times}
\usepackage{epsfig}
\usepackage{graphicx}
\usepackage{amsmath}
\usepackage{amssymb}

\usepackage{amsmath,epsfig}
\usepackage{subfigure}
\usepackage{algorithm}
\usepackage{algorithmic}
\usepackage{booktabs}
\usepackage{multirow}
\usepackage{epstopdf}
\usepackage{marvosym}
\usepackage{array}
\usepackage{fancyhdr}

\usepackage[breaklinks=true,bookmarks=false]{hyperref}

\cvprfinalcopy 

\def\cvprPaperID{****} 

\begin{document}

\title{CFENet: An Accurate and Efficient Single-Shot Object Detector for Autonomous Driving}


\author{Qijie Zhao$^1$, Tao Sheng$^1$, Yongtao Wang$^1$\thanks{Corresponding Author} ,  Feng Ni$^1$, Ling Cai$^2$\\
$^1$Institute of Computer Science and Technology, Peking University\\
$^2$Alibaba AI Lab\\
\tt\small \{zhaoqijie,shengtao,wyt,nifeng\}@pku.edu.cn, cailing.cl@alibaba-inc.com
}

\maketitle

\begin{abstract}

The ability to detect small objects and the speed of  the object detector are very important for the application of autonomous driving, and in this paper, we propose an effective yet efficient one-stage detector, which gained the second place in the Road Object Detection competition of CVPR2018 workshop - Workshop of Autonomous Driving(WAD). The proposed detector inherits the architecture of SSD and introduces a novel Comprehensive Feature Enhancement(CFE) module into it. Experimental results on this competition dataset as well as the MSCOCO dataset demonstrate that the proposed detector (named CFENet) performs much better than the original SSD and the state-of-the-art method RefineDet especially for small objects, while keeping high efficiency close to the original SSD. Specifically, the single scale version of the proposed detector can run at the speed of 21 fps, while the multi-scale version with larger input size achieves the mAP 29.69, ranking second on the leaderboard.
\end{abstract}

\section{Introduction}
For an autonomous driving car, visual perception unit is of great significance \cite{ShiALY17,HouZZZ17} to sense the surrounding scenes \cite{ChenZLZ18,UcarDG17,YeFL16}, and object detector is the core of this unit. An adequate object detector for the application of autonomous driving should be effective and efficient enough, and has strong ability to detect small objects. As shown in Fig 1, there are large portion of small objects in the autonomous driving scenes, thus the ability of detecting small objects is very important in these scenes. Specifically, detecting small objects like traffic lights and traffic signs is crucial to driving planning and decisions, and finding faraway objects appearing small in images is helpful to early make plan for avoid the potential dangers. Besides, detection speed is another important factor \cite{WuIJK17}, since real-time object detection can help driverless cars avoid obstacles in time.


Recently, deep neural network based methods achieve encouraging results for general object detection problem. The state-of-the-art methods for general object detection can be briefly categorized into one-stage methods (e. g., YOLO  \cite{RedmonDGF16}, SSD \cite{LiuAESRFB16}, Retinanet \cite{LinGGHD17}), RefineDet \cite{abs-1711-06897}, and two-stage methods (e.g., Fast/Faster R-CNN \cite{RenHGS15}, FPN \cite{LinDGHHB17}, Mask R-CNN \cite{HeGDG17}). Generally speaking, two-stage methods usually have better detection performance while one-stage methods are more efficient. In this work, we focus on the one-stage detector, due to requirement about the detection speed in the autonomous driving scenes. 
\begin{figure}
\begin{center}
\includegraphics[width=\linewidth]{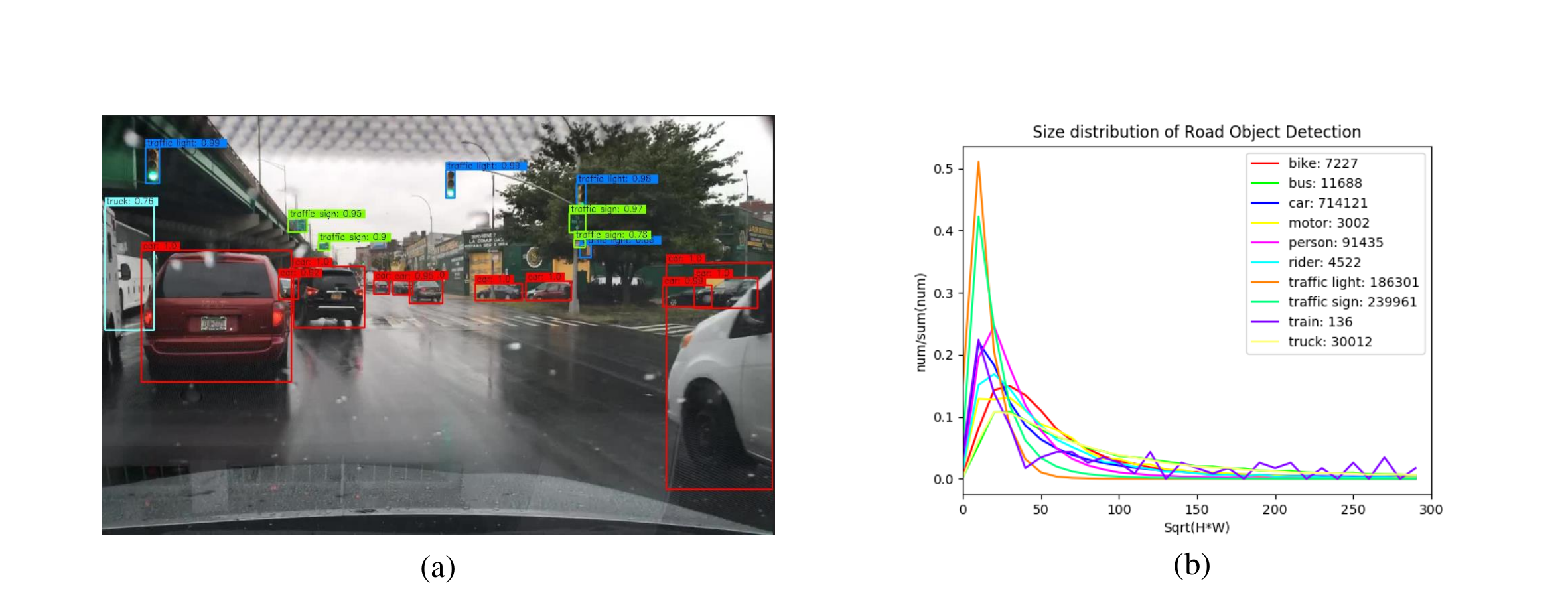}
\end{center}
      \caption{(a) An example detection result of the proposed CFENet on the BDD testing dataset. (b) The distribution of object size of BDD training dataset for each category.}
\label{fig:cfe}
\end{figure}
YOLO \cite{RedmonDGF16} and SSD \cite{LiuAESRFB16} are two representative one-stage detectors. YOLO adopts a relative simple architecture thus very efficient, but cannot deal with dense objects or objects with large scale variants. As for SSD, it could detect objects with different size from multi-scale feature maps. Moreover, SSD uses anchor strategy to detect dense objects. Therefore, it achieves a pretty detection performance. In addition, SSD with input size of 512$\times$512 can achieve the speed of more than 20 FPS on the graphics processing unit(GPU) such as Titan XP. Due to the above advantages, SSD becomes a very practical object detector in industry, which has been widely used for many tasks. However, its performance on small objects is not good. For example, on the test-dev of MSCOCO \cite{LinMBHPRDZ14}, the average precision(AP) of small objects of SSD is only 10.9\%, and the average recall(AR) is only 16.5\%. The major reason is that it uses shallow feature map to detect small objects, which doesn?t contain rich high-level semantic information thus not discriminative enough for classification. The newly proposed RefineDet \cite{abs-1711-06897}, has tried to solve this problem. As shown in Fig 2.b, RefineDet uses an Encode-Decode \cite{FuLRTB17} structure to deepen the network and upsample feature maps so that large-scale feature maps can also learn deeper semantic information. On the other hand, RefineDet uses the idea of cascade regression like Faster-RCNN \cite{RenHGS15}, applying the Encode part to firstly regress coarse positions of the targets, and then use the Decode part to regress out a more accurate position on the basis of the previous step. On MSCOCO test-dev, it gets the average precision of 16.3\% and average recall of 29.3\% for small objects. Also, RefineDet with VGG backbone could performs with high efficiency. Although this result is significantly better than SSD, there is still much room for performance improvement on this dataset.

\begin{figure}
\begin{center}
\includegraphics[width=\linewidth]{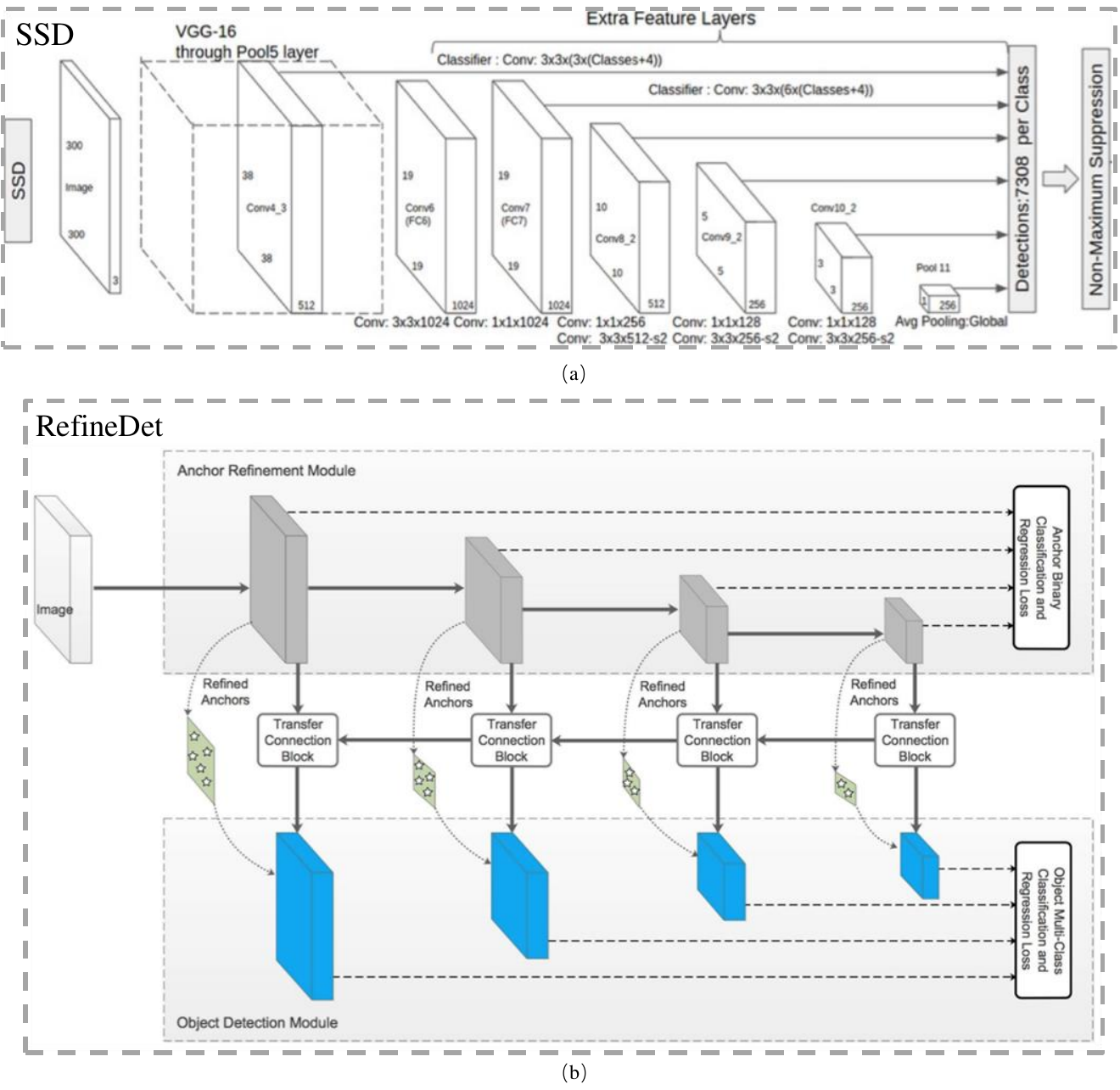}
\end{center}
      \caption{Architectures of the two one-stage object detectors, SSD and RefineDet. (a) SSD with input-size 300$\times$300. (b) RefineDet with input-size 320$\times$320.}
\label{fig:cfe}
\end{figure}

\begin{figure*}[ht]
\begin{center}
\includegraphics[width=\linewidth]{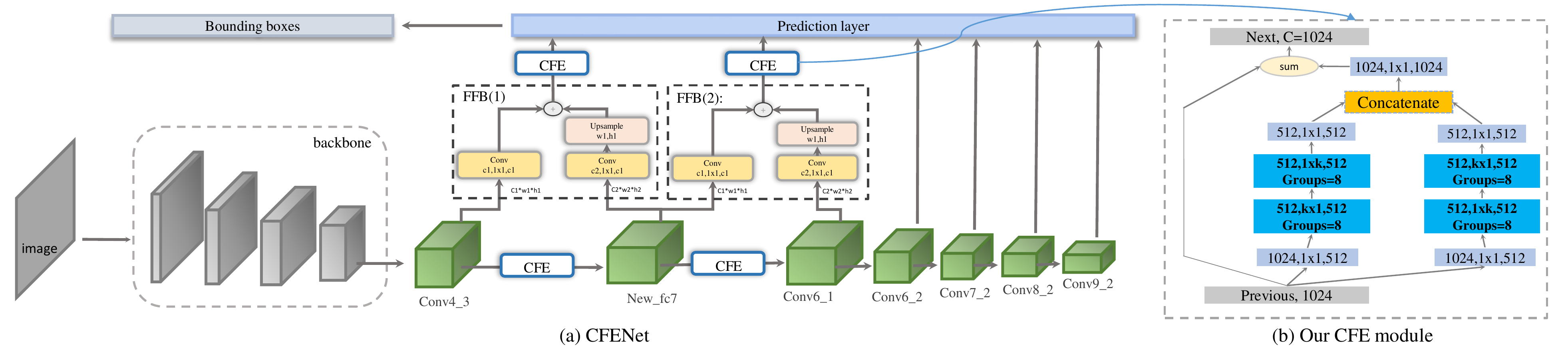}
\end{center}
      \caption{The architecture of CFENet and its novel module CFE. (a) The topology structure of CFENet with input-size 300*300. (b) The layer settings of CFE module, which each box represents a conv+bn+relu group. }
\label{fig:cfe}
\end{figure*}

To address the above mentioned problems, we build a lightweight but effective one-stage network architecture namely CFENet for object detection in autonomous driving scenes. CFENet inherits the architecture of SSD and improves the detection accuracy (especially for the small objects), at the expense of only a very small amount of inference time. The experiment results on MSCOCO show that CFENet performs with higher detection accuracy than the state-of-the-art one-stage detector RefineDet. Moreover, CFENet also significantly outperforms RefineDet for detecting small objects. In detail, CFENet gets 34.8 mAP totally and 18.3 mAP on small objects, exceeding RefineDet by 1.8 points and 2.2 points respectively. On the test set of Road Object Detection task of Berkeley DeepDrive(BDD) \cite{abs-1805-04687}, CFENet800 ranked second, under the evaluating threshold of IoU=0.7.

\section{Architecture of SSD}
Here, we briefly review the most widely used one-stage detector SSD \cite{LiuAESRFB16}, which is the basis of the proposed method CFENet.

As illustrated in Fig 2.a, SSD is a fully convolutional network with a feature pyramid structure. Note that the backbone-inside layer Conv4\_3 is adopted for detecting objects of smallest size, the deeper layers are used to detect relative bigger objects. The range of the anchor size corresponding to each feature map is determined according to the object scale distributions on the training dataset. For anchor matching, it begins by matching each ground truth box to the default box with the best
jaccard overlap, then match default boxes to any ground truth with jaccard overlap higher than a threshold (0.5). Although SSD can alleviate the problems arising from object scale variation, it has limitation to detect small objects. The major reason is that it uses the feature of Conv4\_3 to detect small objects, which is relatively shallow and does not contain rich high-level semantic information. Hence, in this work, we attempt to improve SSD by enhancing the feature for detecting small objects.

\section{CFENet}
As shown in Fig 3.a, CFENet assembles four Comprehensive Feature Enhancement(CFE) modules and two Feature Fusion Blocks(FFB) into original SSD. These additional modules are simple, which can be easily assembled into conventional detection networks. The inner structure of CFE is shown in Fig 3.b, which consists of two similar branches. For example, in the left branch, we use k$\times$k Conv followed by 1$\times$1 Conv \cite{LinCY13} for learning more non-linear relations and broadening the receptive field. meanwhile we factorize the k$\times$k Conv into a 1$\times$k and a k$\times$1 Conv layers for keeping receptive field as well as saving the inference time of CFENet. The difference of the other branch is to reverse the group of 1$\times$k and k$\times$1 conv layers. The CFE module is designed to enhance the shallow features of SSD for detecting small objects, which is actually motivated from multiple existing modules like Inception module \cite{SzegedyVISW16}, Xception module \cite{Chollet17}, Large separable module \cite{abs-1711-07264} and ResNeXt block \cite{XieGDTH17}.

Based on CFE module, we propose a novel one-stage detector CFENet which is more effective for detecting small objects. To be more specific, we first assemble two CFEs between the Conv layers of Conv4\_3 and Fc\_7 and the Conv layers between Fc\_7 and Conv6\_2 respectively. In addition, we connect another two separate CFEs to Conv4\_3 and Fc\_7 detection branches respectively. Because these two layers are relatively shallow, its learned features are still not good for the latter recognition process, we use CFE modules to enhance Conv4\_3 and Fc\_7 features. Step forward, feature fusion strategy always contribute for learning better features that combining advantages from the original features \cite{LongSD15,RonnebergerFB15}. We applied this method in CFENet, too. In detail, generating the new Conv4\_3 and Fc\_7 by feature fusion with the help of two FFBs. We set k=7 for CFE modules in experiment section.

The assembled CFEs could also be placed at other candidate positions, more CFEs will bring more improvement to the original network. Considering the tradeoff between the improved accuracy and increased inference time, we have experimented and select the version shown in Fig.3 finally.

\section{Experiments}
The experiments are conducted on MSCOCO and BDD datasets. We compare SSD, RefineDet and CFENet on MSCOCO to evaluate their performance on overall accuracy and small-objects accuracy. Then we show experiments on BDD dataset of the WAD workshop. In both experiments, the evaluation metric is mean Average Precision(mAP) among all categories. The backbone of CFENet is VGG-16 \cite{SimonyanZ14a}. It's worth noting that, for fair comparison, the three detectors use the same backbone.

\subsection{Experiments on MSCOCO}
MSCOCO is a large dataset with 80 object categories. We use the union of 80k training images and a 35k subset of validation images(trainval35k) to train our model as in \cite{abs-1711-06897,LiuAESRFB16}, and report results from test-dev evaluation server. 

For fair comparison, we report results of single-scale version for each detector. As shown in Tab.1, for both the input size of 300x300 and 512x512, the improvements of CFENet is significant. Notably, CFENet gains mAP of 34.8, achieves state-of-the-art result for one-stage detectors with VGG-16 backbone. Moreover, it also gets AP of 18.5 on small objects, which is the best result for input-size of 512$\times$512. For all scales(small, medium and large), CFENet outperforms RefineDet at least 1 point, which demonstrates that CFENet is a more effective one-stage detector.

\begin{table}[htb]
       \scriptsize
	\caption{Comparison of detection accuracy in terms of mAP on \textit{MS COCO} test-dev set.}
	\label{tab:cocoresults}
	\centering
	\begin{tabular}{p{1.5cm}<{\centering}|p{0.4cm}<{\centering}|p{0.5cm}<{\centering}p{0.5cm}<{\centering}p{0.5cm}<{\centering}|p{0.5cm}<{\centering}p{0.5cm}<{\centering}p{0.5cm}<{\centering}}
		\hline
		\toprule
		\multirow{2}{*}{Method} & \multirow{2}{*}{Size} & \multicolumn{3}{c|}{Avg. Precision, IoU:} & \multicolumn{3}{c}{Avg. Precision, Area:}  \\
	     & & 0.5:0.95 & 0.5 & 0.75 & S & M & L \\
		\hline
		SSD \cite{LiuAESRFB16}  & 300& 25.1 & 43.1 & 25.8 & 6.6 & 25.9 & 41.4 \\
	
		RefineDet \cite{abs-1711-06897}  & 320& 29.4 & 49.2 & \textbf{31.3} & 10.0 & 32.0 & 44.4 \\
		CFENet & 300& \textbf{30.2}& \textbf{50.5}& \textbf{31.3} & \textbf{12.7} & \textbf{32.7}&\textbf{46.6}\\
		\hline
		SSD \cite{LiuAESRFB16}  & 512& 28.8 & 48.5 & 30.3 & 10.9 & 31.8 & 43.5 \\
		RefineDet \cite{abs-1711-06897}  &512& 33.0 & 54.5 & 35.5 & 16.3 & 36.3 & 44.3 \\
		CFENet & 512 & \textbf{34.8} & \textbf{56.3} & \textbf{36.7} & \textbf{18.5} & \textbf{38.4}& \textbf{47.4}\\
		\bottomrule
	\end{tabular}
	\label{tab:Margin_settings}
\end{table}

\begin{figure*}[t]
\begin{center}
\includegraphics[width=\linewidth]{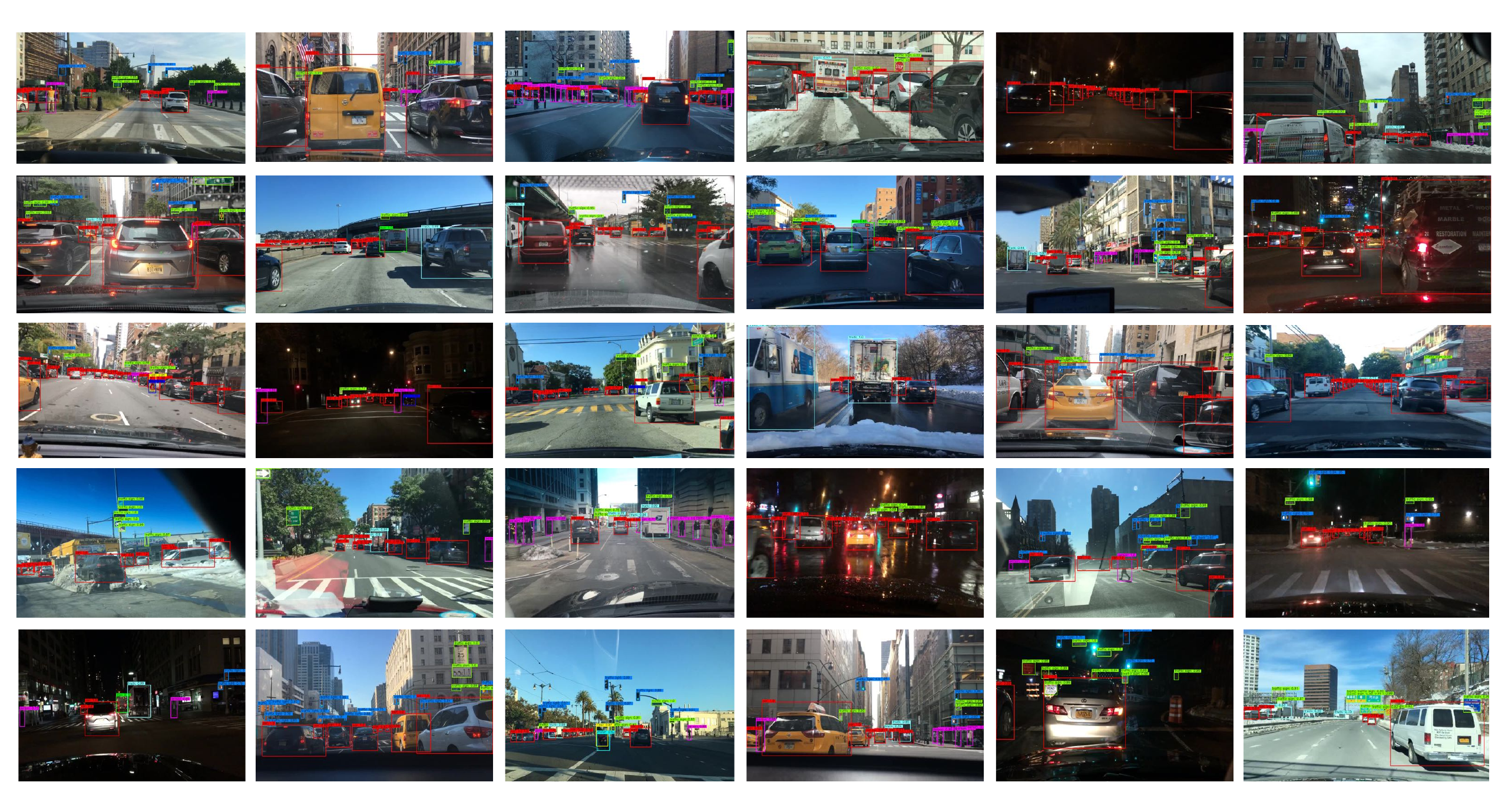}
\end{center}
      \caption{Visualization of some detection results on Berkeley DeepDrive.}
\label{fig:cfe}
\end{figure*}

\textbf{Ablation Study.} To evaluate the contribution of different components of CFENet, we further construct 3 variants and conduct ablation studies to evaluate them. It should be point out that, the results are obtained on minival set of MSCOCO to save time. 

The first step is to validate the effect of CFE module, We choose a channel-broadened Inception module as comparison, assembling the two kinds of modules on SSD at top positions shown in Fig 3.a, and the first three columns of Tab 2 illustrate that the CFE module has a higher promotion(from 28.8 to 31.7, bigger than 30.3). Second, we further insert two CFEs at bottom positions shown in Fig 3.a, the mAP is then increased to 33.9, which shows the effectiveness of the added two CFEs for enhancing the overall features. Finally, after fusing features by two FFBs, mAP rises to 34.8, so that it has demonstrated the effectiveness of FFB. These experiments have proved the importance of each component of CFENet.

\begin{table}[htb]
\centering
\caption{Ablation study of CFENet on MSCOCO.}
\label{tab:SurveyDetails}
\begin{tabular}{cccccc}
\hline
+2 Incep(T) & {} &${\surd}$ &{}& {} & {}  \\
+2 CFE(T) &{}& {} & ${\surd}$ & ${\surd}$ & ${\surd}$  \\
+2 CFE(B) &  {}&{} & {} & ${\surd}$ & ${\surd}$  \\
+2 FFB & {}& {} & {} & {} & ${\surd}$  \\
\hline
mAP & 28.8& 30.3 & 31.7 & 33.9 & \textbf{34.8}  \\ \hline

\end{tabular}
\label{tab:Margin_settings}
\end{table}

\subsection{Experiments on Berkeley DeepDrive(BDD).}
BDD is a well annotated dataset that includes road object detection, instance segmentation, driveable area segmentation and lane markings detection annotations. The road object detection contains 2D bounding boxes annotated on 100,000 images for bus, traffic light, traffic sign, person, bike, truck, motor, car, train, and rider, 10 categories in total. The split ratio of training, validation and testing set is 7:1:2. The evaluated IoU threshold is 0.7 on testing leaderboard.

First, we compare the efficiency of fast version CFENet512 and RefineDet512, i.e., both with single-scale inference strategy, and the experimental results show that they could both run faster than 20 FPS. Second, we compare the accuracy of both detectors. As shown in Tab 3, CFENet512 achieves higher mAP than RefineDet. Specially, for evaluating performance on small objects, we also compare the average accuracy of both detectors on category \textit{traffic light} and \textit{traffic sign}(denoted by \textit{S-mAP} in Tab.3). Obviously, CFENet outperforms RefineDet for detecting such kinds of small objects. 

\begin{table}[htb]
  \caption{Comparison between RefineDet512 and CFENet512 on BDD.}
  \label{tab:detracresults}
  \begin{tabular}{p{1.3cm}<{\centering}|p{1.4cm}<{\centering}|p{0.8cm}<{\centering}|p{1.1cm}<{\centering}|p{1.6cm}<{\centering}}
    \toprule
    Method & Input size &FPS& mAP(\%) & S-mAP(\%)\\
    \midrule
   RefineDet&$512\times512$&\textbf{22.3}&17.4& 13.1\\
   CFENet&$512\times512$&21.0&\textbf{19.1}&\textbf{15.4}\\
\hline
\end{tabular}
\end{table}

Due to the limitations of time and computational resources, we just adopt VGG-16 backbone in this competition, without using more powerful networks, such as ResNet \cite{HeZRS16} and DPN \cite{ChenLXJYF17} backbone. To get a better result on the leaderboard in this competition, we enlarge the input size to 800$\times$800, and boost mAP about 3.2 points. With the optimized efficiency of PyTorch v0.4+, the VGG-CFENet800 could still realize 20+fps with hard-NMS. Furthermore, we adopt multi-scale inference strategy to help detect small objects more accurately, achieving the mAP result of 29.69 on the leaderboard.

\begin{table}[htb]
\centering
 \scriptsize
  \caption{Detailed Results of CFENet800 and CFENet800 - MS on BDD.}
  \label{tab:detracresults}
  \begin{tabular}{p{2.3cm}<{\centering} p{2cm}<{\centering} p{2.5cm}<{\centering}}
     \toprule
    Category name & CFENet800 & CFENet800 - MS \\
    \midrule
   Bike&14.58&20.51\\
   Bus&39.59&50.43\\
   Car&45.20&51.29\\
   Motor&11.48&16.73\\
   Person&17.62&29.08\\
   Rider&12.73&23.86\\
   Traffic light&8.67&15.27\\
   Traffic sign&28.29&37.54\\
   Train&0&0\\
   Truck&45.28&52.23\\\hline
   \textbf{mean}&22.34 & \textbf{29.69}\\
\hline

\end{tabular}

\end{table}

The detailed accuracy of each category is listed in Tab.4, CFENet performs well among most of them. Vehicles like car, bus and truck are easier to detect because they have enough training samples while class \textit{train} is difficult due to lackness of positive samples in training set. 
In addition, we have visualized a number of detection results in Fig.4.

\section{Conclusion}
In this WAD Berkeley DeepDrive(BDD) Road Object Detection challenge, we have proposed an effective one-stage architecture, CFENet, based on SSD and a novel Comprehensive Feature Enhancement(CFE) module. The multi-scale version of CFENet800 achieves 29.69 of mAP on the final testing leaderboard, ranking second on the testing leaderboard. Moreover, experimental results on MSCOCO and BDD reveal that CFENet has significantly outperformed the original SSD and state-of-the-art one-stage object detector RefineDet, especially for detecting small objects. In addition, the single scale version of CFENet512 can achieve real-time speed, i.e., 21fps.  These advantages demonstrate that CFENet is more suitable for the application of autonomous driving.


{\small
\bibliographystyle{ieee}
\bibliography{egpaper_final}

\begin{thebibliography}{10}\itemsep=-1pt

\bibitem{ChenLXJYF17}
Y.~Chen, J.~Li, H.~Xiao, X.~Jin, S.~Yan, and J.~Feng.
\newblock Dual path networks.
\newblock In {\em Advances in Neural Information Processing Systems 30: Annual
  Conference on Neural Information Processing Systems 2017, 4-9 December 2017,
  Long Beach, CA, {USA}}, pages 4470--4478, 2017.

\bibitem{ChenZLZ18}
Y.~Chen, D.~Zhao, L.~Lv, and Q.~Zhang.
\newblock Multi-task learning for dangerous object detection in autonomous
  driving.
\newblock {\em Inf. Sci.}, 432:559--571, 2018.

\bibitem{Chollet17}
F.~Chollet.
\newblock Xception: Deep learning with depthwise separable convolutions.
\newblock In {\em 2017 {IEEE} Conference on Computer Vision and Pattern
  Recognition, {CVPR} 2017, Honolulu, HI, USA, July 21-26, 2017}, pages
  1800--1807, 2017.

\bibitem{FuLRTB17}
C.~Fu, W.~Liu, A.~Ranga, A.~Tyagi, and A.~C. Berg.
\newblock {DSSD} : Deconvolutional single shot detector.
\newblock {\em CoRR}, abs/1701.06659, 2017.

\bibitem{HeGDG17}
K.~He, G.~Gkioxari, P.~Doll{\'{a}}r, and R.~B. Girshick.
\newblock Mask {R-CNN}.
\newblock In {\em {IEEE} International Conference on Computer Vision, {ICCV}
  2017, Venice, Italy, October 22-29, 2017}, pages 2980--2988, 2017.

\bibitem{HeZRS16}
K.~He, X.~Zhang, S.~Ren, and J.~Sun.
\newblock Deep residual learning for image recognition.
\newblock In {\em 2016 {IEEE} Conference on Computer Vision and Pattern
  Recognition, {CVPR} 2016, Las Vegas, NV, USA, June 27-30, 2016}, pages
  770--778, 2016.

\bibitem{HouZZZ17}
Y.~Hou, H.~Zhang, S.~Zhou, and H.~Zou.
\newblock Efficient convnet feature extraction with multiple roi pooling for
  landmark-based visual localization of autonomous vehicles.
\newblock {\em Mobile Information Systems}, 2017:8104386:1--8104386:14, 2017.

\bibitem{abs-1711-07264}
Z.~Li, C.~Peng, G.~Yu, X.~Zhang, Y.~Deng, and J.~Sun.
\newblock Light-head {R-CNN:} in defense of two-stage object detector.
\newblock {\em CoRR}, abs/1711.07264, 2017.

\bibitem{LinCY13}
M.~Lin, Q.~Chen, and S.~Yan.
\newblock Network in network.
\newblock {\em CoRR}, abs/1312.4400, 2013.

\bibitem{LinDGHHB17}
T.~Lin, P.~Doll{\'{a}}r, R.~B. Girshick, K.~He, B.~Hariharan, and S.~J.
  Belongie.
\newblock Feature pyramid networks for object detection.
\newblock In {\em 2017 {IEEE} Conference on Computer Vision and Pattern
  Recognition, {CVPR} 2017, Honolulu, HI, USA, July 21-26, 2017}, pages
  936--944, 2017.

\bibitem{LinGGHD17}
T.~Lin, P.~Goyal, R.~B. Girshick, K.~He, and P.~Doll{\'{a}}r.
\newblock Focal loss for dense object detection.
\newblock In {\em {IEEE} International Conference on Computer Vision, {ICCV}
  2017, Venice, Italy, October 22-29, 2017}, pages 2999--3007, 2017.

\bibitem{LinMBHPRDZ14}
T.~Lin, M.~Maire, S.~J. Belongie, J.~Hays, P.~Perona, D.~Ramanan,
  P.~Doll{\'{a}}r, and C.~L. Zitnick.
\newblock Microsoft {COCO:} common objects in context.
\newblock In {\em Computer Vision - {ECCV} 2014 - 13th European Conference,
  Zurich, Switzerland, September 6-12, 2014, Proceedings, Part {V}}, pages
  740--755, 2014.

\bibitem{LiuAESRFB16}
W.~Liu, D.~Anguelov, D.~Erhan, C.~Szegedy, S.~E. Reed, C.~Fu, and A.~C. Berg.
\newblock {SSD:} single shot multibox detector.
\newblock In {\em Computer Vision - {ECCV} 2016 - 14th European Conference,
  Amsterdam, The Netherlands, October 11-14, 2016, Proceedings, Part {I}},
  pages 21--37, 2016.

\bibitem{LongSD15}
J.~Long, E.~Shelhamer, and T.~Darrell.
\newblock Fully convolutional networks for semantic segmentation.
\newblock In {\em {IEEE} Conference on Computer Vision and Pattern Recognition,
  {CVPR} 2015, Boston, MA, USA, June 7-12, 2015}, pages 3431--3440, 2015.

\bibitem{RedmonDGF16}
J.~Redmon, S.~K. Divvala, R.~B. Girshick, and A.~Farhadi.
\newblock You only look once: Unified, real-time object detection.
\newblock In {\em 2016 {IEEE} Conference on Computer Vision and Pattern
  Recognition, {CVPR} 2016, Las Vegas, NV, USA, June 27-30, 2016}, pages
  779--788, 2016.

\bibitem{RenHGS15}
S.~Ren, K.~He, R.~B. Girshick, and J.~Sun.
\newblock Faster {R-CNN:} towards real-time object detection with region
  proposal networks.
\newblock In {\em Advances in Neural Information Processing Systems 28: Annual
  Conference on Neural Information Processing Systems 2015, December 7-12,
  2015, Montreal, Quebec, Canada}, pages 91--99, 2015.

\bibitem{RonnebergerFB15}
O.~Ronneberger, P.~Fischer, and T.~Brox.
\newblock U-net: Convolutional networks for biomedical image segmentation.
\newblock In {\em Medical Image Computing and Computer-Assisted Intervention -
  {MICCAI} 2015 - 18th International Conference Munich, Germany, October 5 - 9,
  2015, Proceedings, Part {III}}, pages 234--241, 2015.

\bibitem{ShiALY17}
W.~Shi, M.~B. Alawieh, X.~Li, and H.~Yu.
\newblock Algorithm and hardware implementation for visual perception system in
  autonomous vehicle: {A} survey.
\newblock {\em Integration}, 59:148--156, 2017.

\bibitem{SimonyanZ14a}
K.~Simonyan and A.~Zisserman.
\newblock Very deep convolutional networks for large-scale image recognition.
\newblock {\em CoRR}, abs/1409.1556, 2014.

\bibitem{SzegedyVISW16}
C.~Szegedy, V.~Vanhoucke, S.~Ioffe, J.~Shlens, and Z.~Wojna.
\newblock Rethinking the inception architecture for computer vision.
\newblock In {\em 2016 {IEEE} Conference on Computer Vision and Pattern
  Recognition, {CVPR} 2016, Las Vegas, NV, USA, June 27-30, 2016}, pages
  2818--2826, 2016.

\bibitem{UcarDG17}
A.~U{\c{c}}ar, Y.~Demir, and C.~G{\"{u}}zelis.
\newblock Object recognition and detection with deep learning for autonomous
  driving applications.
\newblock {\em Simulation}, 93(9):759--769, 2017.

\bibitem{WuIJK17}
B.~Wu, F.~N. Iandola, P.~H. Jin, and K.~Keutzer.
\newblock Squeezedet: Unified, small, low power fully convolutional neural
  networks for real-time object detection for autonomous driving.
\newblock In {\em 2017 {IEEE} Conference on Computer Vision and Pattern
  Recognition Workshops, {CVPR} Workshops, Honolulu, HI, USA, July 21-26,
  2017}, pages 446--454, 2017.

\bibitem{XieGDTH17}
S.~Xie, R.~B. Girshick, P.~Doll{\'{a}}r, Z.~Tu, and K.~He.
\newblock Aggregated residual transformations for deep neural networks.
\newblock In {\em 2017 {IEEE} Conference on Computer Vision and Pattern
  Recognition, {CVPR} 2017, Honolulu, HI, USA, July 21-26, 2017}, pages
  5987--5995, 2017.

\bibitem{YeFL16}
Y.~Ye, L.~Fu, and B.~Li.
\newblock Object detection and tracking using multi-layer laser for autonomous
  urban driving.
\newblock In {\em 19th {IEEE} International Conference on Intelligent
  Transportation Systems, {ITSC} 2016, Rio de Janeiro, Brazil, November 1-4,
  2016}, pages 259--264, 2016.

\bibitem{abs-1805-04687}
F.~Yu, W.~Xian, Y.~Chen, F.~Liu, M.~Liao, V.~Madhavan, and T.~Darrell.
\newblock {BDD100K:} {A} diverse driving video database with scalable
  annotation tooling.
\newblock {\em CoRR}, abs/1805.04687, 2018.

\bibitem{abs-1711-06897}
S.~Zhang, L.~Wen, X.~Bian, Z.~Lei, and S.~Z. Li.
\newblock Single-shot refinement neural network for object detection.
\newblock {\em CoRR}, abs/1711.06897, 2017.

\end{thebibliography}
}

\end{document}